\documentclass[lettersize,journal]{IEEEtran}
\usepackage{amsmath,amsfonts}
\usepackage{algorithmic}
\usepackage{algorithm}
\usepackage{array}
\usepackage[caption=false,font=normalsize,labelfont=sf,textfont=sf]{subfig}
\usepackage{textcomp}
\usepackage{stfloats}
\usepackage{url}
\usepackage{verbatim}
\usepackage{graphicx}
\usepackage{cite}
\usepackage{graphicx}
\usepackage{multirow}
\usepackage{booktabs}
\usepackage{bbm}
\usepackage{color}
\newcommand\R[1]{\textcolor{black}{#1}}
\newcommand{\RomanNumeralCaps}[1]
    {\MakeUppercase{\romannumeral #1}}
\begin{document}

\title{Semi-Supervised Crowd Counting with Contextual Modeling: Facilitating Holistic Understanding of Crowd Scenes}

\author{Yifei Qian, Xiaopeng Hong,
\IEEEmembership{Senior Member, IEEE}, Zhongliang Guo, Ognjen~Arandjelovi\'c, Carl R.Donovan
\thanks{This work is funded by the National Natural Science Foundation of China (62076195, 62376070, 62206271) and the Fundamental Research Funds for the Central Universities (AUGA5710011522). Zhongliang Guo acknowledges the financial support through the China Scholarship Council – University of St Andrews Scholarship (Grant No.202208060113). (\textit{Corresponding author: Xiaopeng Hong.})}
\thanks{Yifei Qian and Carl R.Donovan are with the School of Maths and Statistics, University of St Andrews, KY16 9AJ St Andrews, U.K.\linebreak Email:\{yq1, crd2\}@st-andrews.ac.uk}
\thanks{Xiaopeng Hong is with the Harbin Institute of Technology, Harbin, Heilongjiang, 150001, China. E-mail: hongxiaopeng@ieee.org.}
\thanks{Zhongliang Guo and Ognjen~Arandjelovi\'c are with the School of Computer Science, University of St Andrews, KY16 9AJ St Andrews, U.K.\linebreak Email:\{zg34, oa7\}@st-andrews.ac.uk}
}

\markboth{Journal of \LaTeX\ Class Files,~Vol.~14, No.~8, August~2015}%
{Shell \MakeLowercase{\textit{et al.}}: Bare Demo of IEEEtran.cls for IEEE Journals}

\makeatletter
\def\ps@IEEEtitlepagestyle{%
  \def\@oddfoot{\mycopyrightnotice}%
  \def\@oddhead{\hbox{}\@IEEEheaderstyle\leftmark\hfil\thepage}\relax
  \def\@evenhead{\@IEEEheaderstyle\thepage\hfil\leftmark\hbox{}}\relax
  \def\@evenfoot{}%
}
\def\mycopyrightnotice{%
  \begin{minipage}{\textwidth}
  \centering \scriptsize
  Copyright~\copyright~20xx IEEE. Personal use of this material is permitted. Permission from IEEE must be obtained for all other uses, in any current or future media, including\\reprinting/republishing this material for advertising or promotional purposes, creating new collective works, for resale or redistribution to servers or lists, or reuse of any copyrighted component of this work in other works by sending a request to pubs-permissions@ieee.org.
  \end{minipage}
}
\makeatother



\maketitle

\begin{abstract}
To alleviate the heavy annotation burden for training a reliable crowd counting model and thus make the model more practicable and accurate by being able to benefit from more data, this paper presents a new semi-supervised method based on the mean teacher framework. When there is a scarcity of labeled data available, the model is prone to overfit local patches. Within such contexts, the conventional approach of solely improving the accuracy of local patch predictions through unlabeled data proves inadequate. Consequently, we propose a more nuanced approach: fostering the model's intrinsic \R{`}subitizing' capability. This ability allows the model to accurately estimate the count in regions by leveraging its understanding of the crowd scenes, mirroring the human cognitive process. To achieve this goal, we apply masking on unlabeled data, guiding the model to make predictions for these masked patches based on the holistic cues. Furthermore, to help with feature learning, herein we incorporate a fine-grained density classification task. Our method is general and applicable to most existing crowd counting methods as it doesn't have strict structural or loss constraints. In addition, we observe that the model trained with our framework shows \R{strong contextual modeling capabilities, which allows it to make robust predictions even when some local details of patches are lost}. Our method achieves the state-of-the-art performance, surpassing previous approaches by a large margin on challenging benchmarks such as ShanghaiTech A and UCF-QNRF. The code is available at: \url{https://github.com/cha15yq/MRC-Crowd.} 
\end{abstract}


\section{Introduction}
\IEEEPARstart{C}{rowd} counting is a computer vision task that refers to the automated quantification of the number of objects or individuals in an image or video. It has found widespread applications in numerous areas such as crowd management, traffic control, urban planning, and wildlife monitoring~\cite{application, wildlife_monitor}, garnering significant attention. In recent years, the rapid progress in deep learning techniques has accelerated the development of numerous fully supervised methods~\cite{ma2019bayesian, wang2020DMCount,  Lin_cvpr22, Han_2023_ICCV, Shu2023Fre, Zhang20223DCC} to address this task. However, it is important to acknowledge that the efficacy of these methods relies on the availability of a substantial amount of labeled data due to the data-hunger nature of the deep learning algorithm. Despite the ease of acquiring crowd scene images, the process of finely annotating them, e.g.\ by  marking each head with a dot, is time-consuming and demanding. As an illustration, the total annotation process for the NWPU-crowd dataset~\cite{NWPU} costs 3000 human hours, involving the creation of 2.13 million annotations. To reduce this dependence on vast quantities of labeled data, this paper focus on semi-supervised crowd counting which leverages massive unlabeled corpora~\cite{Zhang2022Weak} to improve the performance of a counting model.

Current mainstream semi-supervised learning~(SSL) methods~\cite{Chen2022, Yang2023, wang2024} can broadly be categorized into two groups, namely those using consistency regularization~\cite{MT} and those relying on pseudo-labeling~\cite{pl}, and have predominantly been developed and studied in the context of classification. However, crowd counting is commonly approached as a regression task, where the objective of the model is to predict a density map for an input image~\cite{L2C}. A straightforward challenge in crowd counting is to obtain an uncertainty measure~\cite{Ranjan_2020_ACCV} for its prediction which is typically necessary in SSL for avoiding learning from incorrect labels~\cite{STC-Crowd}. Unlike classification tasks, where the softmax outputs can be directly used for evaluating uncertainty, regression tasks lack an inherent measure of uncertainty due to their deterministic nature. Hence, several approaches have been proposed in literature to handle this problem and adapt existing SSL methods for crowd counting. For instance, SUA~\cite{Meng2021SpatialUS} adapts the consistency-based mean teacher framework~\cite{zhao2022uda, zhao2021mt, zhao2022mmgl, zhao2022act, li2021hierarchical, zhao2019semi} for crowd counting which leverages the Shannon entropy to derive the uncertainty measure for density estimation from the prediction of a binary classification task. Under the same basic framework, STC-Crowd~\cite{STC-Crowd} utilizes Masksembles for evaluating the prediction confidence and places an extra focus on temporal consistency. MTCP~\cite{MTCP} is a pseudo-label based method, which uses an additional confidence branch to measure the credibility of pseudo-label and hence reduces the noise in these labels. Despite their efforts, the performance of these methods remains inadequate, which prompts us to rethink the challenges of semi-supervised crowd counting. 

\begin{figure}[t]
  \centering
  \includegraphics[width=\linewidth]{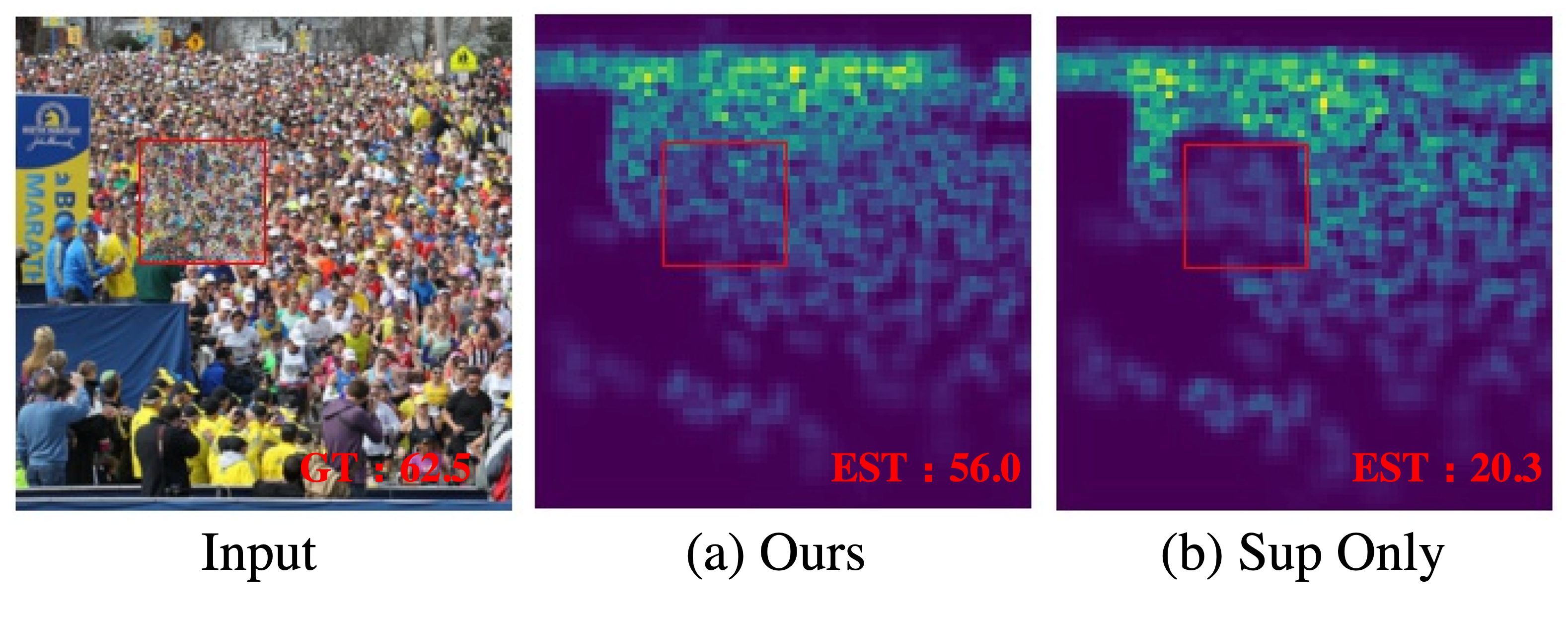}
  \caption{This illustration shows the problem of models excessively relying on the self-information of individual patches when confronted with limited labeled data. On the far left is the input image with a specific region containing Gaussian noise highlighted by a red rectangle. We show the predicted density map from two models, (a) Ours: the model trained with our proposed method under 40\% labeled setting; (b) Sup Only: the same model trained under the same setting but with labeled data only. The count within that region is indicated on the bottom-right of each image. Both models share the same network structure which is detailed in section \RomanNumeralCaps{3} as the \R{`}base network'.}
  \label{fig:problem}
\end{figure}

We begin by investigating the underlying factors contributing to the poor generalization performance within models trained with limited labeled data. Empirically, as shown in Fig~\ref{fig:problem}, we find that the performance of a model trained solely on labeled data degrades significantly when predicting on a noisy patch. 
Therefore, we deduce in the semi-supervised setting, models tend to overfit the limited labeled dataset due to an excessive reliance on the local information of a patch, \R{ which could make the model sensitive to changes in the appearance or arrangement of the crowd, undermining its robustness and reliability. We will further demonstrate this issue in the last part of Section~IV. In addition, it is foreseeable that the model's contextual modeling ability is weakened}. However, existing semi-supervised crowd counting frameworks often overlook this aspect. As mentioned earlier, they primarily focus on constructing a reliable uncertainty measure for the model's predictions on the patch and selecting high-confidence samples from unlabeled data to improve local patch performance. In addition, the effectiveness of the existing methods could also be influenced by the inherent imbalance in crowd density. Since patches with low-density level are prevalent in crowd images, they are more likely to be selected for training the model~\cite{Crest} thereby introducing bias and causing the model to exhibit a tendency of making lower density predictions. Given this context, we propose a novel approach to tackle the semi-supervised crowd counting problem from a distinctive perspective: we foster the model's holistic understanding of the crowd scene with unlabeled images instead of solely improving accuracy on some patches.

We are in part motivated by the observations that humans perform similar counting tasks using an understanding of the scene as a whole~\cite{Jevons1871ThePO}. The cognitive phenomenon known as \R{`}subitizing' describes an ability of human~\cite{Jevons1871ThePO} and even certain animals~\cite{Gross2009NumberBasedVG, Pahl2013}, to accurately and effortlessly estimate the quantity of objects within a small set, usually up to about four items, by a simple glance, without the need for explicit counting. The underlying cognitive mechanism behind subitizing, suggested by the psychological observations~\cite{Jansen2014, Clements1999SubitizingWI, Mandler1982SubitizingAA}, is the brain's remarkable ability to efficiently recognize holistic patterns~\cite{Zhang_2023_CVPR}. However, the significance of contextual modeling has been consistently underestimated in prior semi-supervised crowd counting methods. To the best of our knowledge, existing semi-supervised crowd counting approaches lack an appropriate method to enhance the model's ability towards contextual understanding. Hence, in this work, we propose a general method to teach a model to predict with holistic cues and show the importance of such ability towards the counting performance. To be specific, we propose a simple mean teacher based semi-supervised framework for crowd counting. To enhance the scene comprehension of the model, we provide the student model with unlabeled data that has been partially masked. Meanwhile, the teacher model receives fully-visible images to generate supervision signals for the masked parts. Under this manner, the student model is encouraged to make predictions on invisible patches using holistic cues, resulting in a shift away from excessive reliance on local patterns and instead comprehend the essence of a scene's composition holistically.

Moreover, we notice that the success of semi-supervised learning~\cite{RasmusBHVR15, 8417973} typically relies on exploring the underlying manifold structure of the data distribution. However, the low entropy feature learnt from the regression loss could not benefit the learning of manifold space~\cite{ZhangYMZY23}. In addition, the regression task inherently presents challenges when attempting to explore the manifold space as it lacks explicit mechanisms for learning and modeling the density relationships between different regions. To alleviate this issue, \R{we adopt the joint modeling of regression and classification by introducing a heterogeneous fine-grained density classification task in the model which helps to capture the density relationships and geometric properties of the data manifold.} This auxiliary task is implemented by a two-layer classification head as a plug-in module. Experimental results show the effectiveness of this simple module for boosting the overall performance when learning from unlabeled data. We perform extensive experiments on four challenging crowd counting datasets and achieve the state-of-the-art results. 

In summary, the main contributions of this paper are three-fold:

\begin{itemize}
\item Inspired by the importance of holistic patterns the cognitive phenomenon of subitizing, we propose utilizing unlabeled images to enhance the overall understanding of the scene for counting models, which effectively alleviates the issue of the model overfitting to local details in the semi-supervised problem.

\item We propose a simple semi-supervised crowd counting framework, termed MRC-Crowd. It achieves a better understanding of the crowd scenes by reconstruction of density information for masked patches in unlabeled images. Our framework is highly versatile, requiring only a two-layer classification head as a plug-in to facilitate feature learning, allowing it to be easily applied to a wide range of counting models.

\item We establish new state-of-the-art results across several benchmarks, which further proves the importance of the capability of contextual modeling in counting models. On the challenging UCF-QNRF~\cite{qnrf} dataset in particular, our method achieves an average reduction of 13.2\% in the mean absolute error and 14.8\% in the mean squared error across all three labeling ratios. 
\end{itemize} 

\section{Related Work}
 \begin{figure*}[t]
  \centering
  \includegraphics[width=\linewidth]{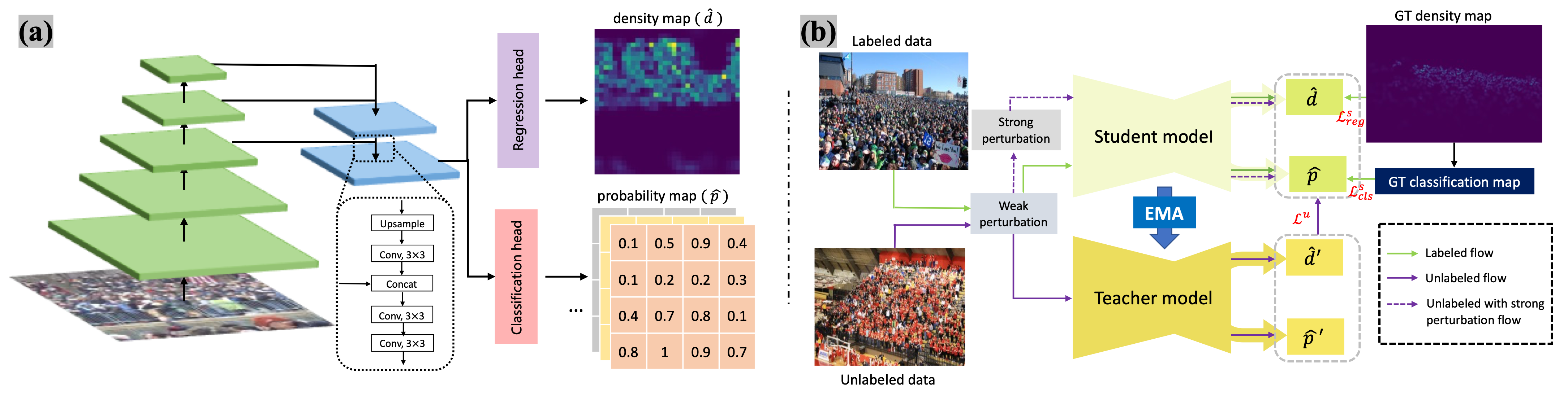}
  \caption{(a) The base network structure adopted in MRC-Crowd. It contains a backbone network, a top-down multi-level feature fusion module, a regression head and a classification head. We adopt VGG-19 as backbone network. (b) The diagram of the overall framework of the proposed MRC-Crowd. The labeled data is used for training the student model by optimizing \(\mathcal{L}^{s}_{reg}\) and \(\mathcal{L}^{s}_{cls}\). The teacher model is updated with the exponential moving average of weights of the student model. The unlabeled data with strong perturbation is fed to the student model while the supervision signals are provided by the predictions of the teacher model on the same data without strong perturbation. Both regression task and the classification task are supervised. The unsupervised learning process is optimized with \(\mathcal{L}^{u}\).}
  \label{fig:framework}
\end{figure*}
\subsection{Fully-supervised crowd counting} In recent years, the development of convolutional neural networks~(CNNs) has driven progress in the field of crowd counting. Many methods~\cite{Gao2019PCCNP, Zhao2020, Zoom, Cao2022, Jiang2020Mask, 9156690, 10256036, Liu_2021_ICCV} have been proposed to improve the counting accuracy in the crowd scenes. For instance, MCNN~\cite{MCNN} employs multiple columns of stacked CNN layers with various receptive fields to handle scale variations in head sizes. CSRNet~\cite{li2018csrnet} leverages dilated convolutions to gain large receptive field, enabling the capture of multi-scale information. These success highlights the importance of leveraging contextual information. Some studies focus on the issue of inaccurate annotation problem in crowd counting~\cite{Wan2023ModelingNA}. BL~\cite{ma2019bayesian} introduces a novel loss function based on Bayes theorem, allowing the model to learn directly from the ground-truth point map. Out of the same purpose, several other works have incorporated optimal transport~\cite{wang2020DMCount, MA2021AAAI, GL, Lin2021DMM} theory into crowd counting. Moreover, to enhance model performance, some approaches have designed to leverage auxiliary tasks such as foreground classification~\cite{ZhaoZZZ19, ShiMS19, shi2023focus}, depth estimation~\cite{Gao2019PCCNP} and perspective estimation~\cite{Zhao2020, Yang2021}. Recently, with the rise of transformer in computer vision, some research have introduced transformers~\cite{sun2021boosting,CUT, Lin_cvpr22, Huang_2023_ICCV, Wu2022trans, Liu_2021_CVPR} into the field of crowd counting to remedy the challenge of capturing long-distance relationship between pixels.

There are also a few studies directly approached crowd counting as a classification problem directly. \R{Liu \textit{et al.}~\cite{Liu2020Bloackwise} introduce a framework with blockwise count level classification.} S-DCNet~\cite{xhp2019SDCNet} leverages both a count classifier and a division decider to transform the open-set counting into the closed-set counting problem. UEPNet~\cite{WangSZWTHWL0W21} proposes an interval partition criteria and the mean count proxy criteria to further improve the performance of purely classification network. However, whatever the approach, achieving a reliable crowd counter typically demands a large annotated dataset, which can be costly to obtain. Therefore, the development of methods that can alleviate the annotation burden while delivering high counting performance is desirable.

\subsection{Semi-supervised crowd counting}
To mitigate the labeling burden, many researchers have proposed to leverage unlabeled images in a semi-supervised manner to improve the performance of a model~\cite{ijcai2023p554, Zhao_2022_BMVC, pahwa20233d, Yang2020Weak, Yinjie2021Count}. These methods can be broadly divided into two groups: direct and indirect. Direct methods try to obtain supervision signals for the regression task from unlabeled images. Typically, an uncertainty measure is constructed for evaluating the quality of regression outputs and then the model is trained with high-confidence samples by leveraging consistency learning~\cite{Meng2021SpatialUS,STC-Crowd, Lin_2023_CVPR} or pseudo-labeling techniques~\cite{MTCP}, resulting in an improvement on the model's performance. However, these methods run the risk of causing the model to overly focus on local information~\cite{guo2023whitebox}, neglecting the global context as they encourage the learning from selected reliable regions. Indirect methods adopt a strategy of designing auxiliary tasks to enhance intermediate feature representations and thereby improving the counting performance. L2R~\cite{Liu2018cvpr} proposes a ranking task to explore the containment relationship within the cropped unlabeled image pairs. IRAST~\cite{liu2020eccv} designs a series of interrelated binary classification tasks and exploits the underlying constraints among them. DACount~\cite{DACount} employs multiple density agents to promote similar feature representations for patches with close density values. Distinct to previous methods that fixated on the accuracy of individual patches, our approach is centered on improving the model's holistic understanding of crowd scenes, resulting in outstanding performance compared to existing methods. Although also based on consistency learning, our method does not necessitate the requirement of an uncertainty measure, which makes our framework more straightforward to implement and easily applicable to various models.

\section{Method}
\subsection{Overview}

\subsubsection{Problem definition} Suppose we have a labeled dataset \(L = \{(x_i^l, D_i^{gt})\}_{i=1}^{N_l}\) consisting of \(N_l\) crowd images, denoted as \(x^l\), along with their corresponding ground-truth density map \(D^{gt}\) generated from the point annotations, and an unlabeled dataset \(U=\{x_i^u\}_{i=1}^{N_u}\) of \(N_u\) samples, where \(N_u \gg N_l\). The objective is to leverage \(U\) to enhance the performance of the crowd counting model and thus surpass what can be achieved by training solely on \(L\). 

\subsubsection{The overall framework} The overall framework of the proposed MRC-Crowd is shown in Fig~\ref{fig:framework}(b). It is based on the traditional mean teacher framework where the student and teacher share the same architecture and the teacher is updated with the expotential moving average (EMA) weights of the student. The base model tested here is given in Fig~\ref{fig:framework}(a). \R{It adopts the joint regression and classification modeling, which leverages a backbone model to extract multi-scale feature and incorporates a top-down feature merging structure to integrate high-level semantic information with low-level details. The resulting feature maps are then processed by both a regression head, and a classification head, facilitating improved feature learning.} Note, the feature merging structure is not necessary and can be removed or replaced with other designs. To be consistent with mainstream fully-supervised models, we set the downsampling ratio of the final feature map to 8. The classification head contains two convolutional layers, each with a kernel size of 1. This design ensures key information is captured in the intermediate feature map that shared with the regression task, while only introduces a very few extra parameters. The regression head consists of three consecutive convolutional layers. The first two layers have a kernel size of 3, while the last layer has a kernel size of 1, which is a common configuration in crowd counting networks.

 \begin{figure}[t]
  \centering
  \includegraphics[width=\linewidth]{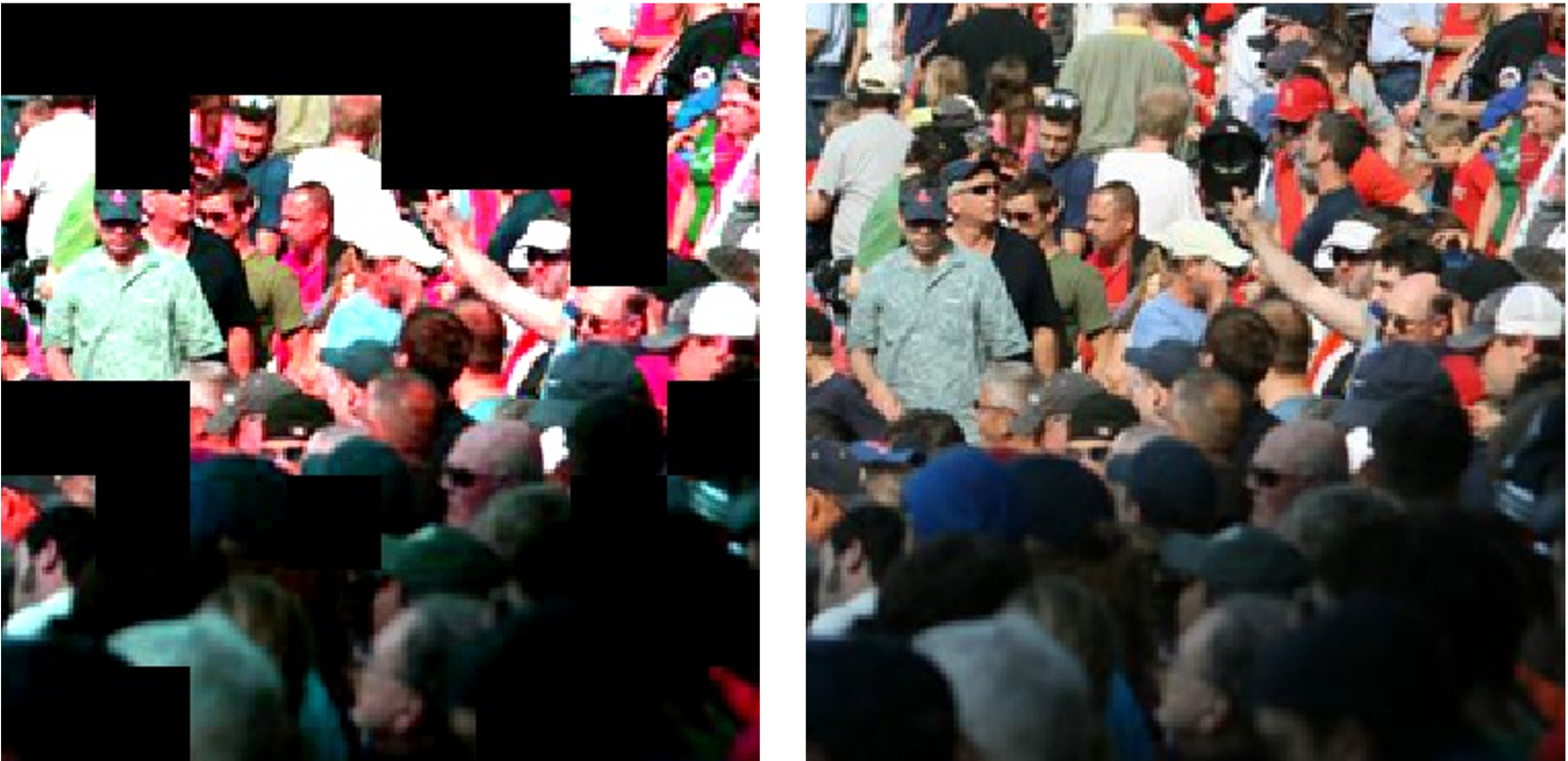}
  \caption{Shown on the left is an example that has undergone strong augmentation (masked patch size$=32$, masking ratio$=0.3$) and on the right the original image.}
  \label{fig:strong_aug}
\end{figure}

The labeled samples are used to train the student model by leveraging the supervised loss, denoted as \(\mathcal{L}^{s}\). This loss consists of two components employed for the density estimation and density classification tasks, respectively. The unlabeled samples are exploited for encouraging model to predict with holistic cues, in order to mitigate the risk of overfitting to local patches. To achieve this goal, we intentionally obscure parts of the image from the student model while still requiring it to make predictions on these masked regions. Apart from masking, we also perform random ColorJitter~\cite{FrenchM22} to avoid model relying on colour statistics to make predictions. The combination of these two techniques is referred as strong perturbation. In Fig~\ref{fig:strong_aug}, the effect of strong perturbation on the input data is illustrated. To achieve supervision signals, we employ the teacher model to predict on the weakly perturbed sample. These weak perturbations involve random horizontal flipping and random scaling with the range of [0.7, 1.3].  \(\mathcal{L}^{u}\) is the unsupervised loss and is used to optimize this process. We will elaborate the details of \(\mathcal{L}^{u}\) and \(\mathcal{L}^{s}\) in the subsequent sections.

The overall loss function can be written as 
\begin{equation}
    \mathcal{L} =  \mathcal{L}^{s} + \lambda_u \cdot\mathcal{L}^{u},
\end{equation}
where \(\lambda_u\) is a weight factor for the unsupervised loss.

\subsection{Supervised Loss}
 The supervised loss has two components. The first is a regression loss for the density estimation task. There are many existing loss functions designed for the crowd counting task. In contrast to some existing semi-supervised crowd counting frameworks~\cite{Lin_2023_CVPR, li2023calibrating} that impose strict requirements on the regression loss function used for training, our proposed framework does not have such constraints, which is also one of our advantages. Here we directly adopt the loss proposed in CUT~\cite{CUT} as the regression loss, which leverages the prior knowledge to improve  the overall accuracy on dense regions. It can be expressed as

\begin{equation}
\begin{aligned}
    \mathcal{L}^{s}_{reg} =&\dfrac{1}{J} \sum_{j=1}^{J}{\left(1 - SSIM\left(P_j\left(\hat{y}\odot M^{gt}\right), P_j\left(y^{gt}\odot M^{gt}\right)\right)\right)} \\
        + &\alpha \cdot\mathcal{L}_{TV}\left(\hat{y}, y^{gt}\right),
\end{aligned}
\end{equation}
where \(SSIM\) represents the structural similarity index measure~\cite{ssim} and \(P_j\) denotes the average pooling operation that downsamples an image to \(\frac{1}{2^{j-1}}\) size. The symbol \(\odot\) denotes the Hadamard multiplication. \(M^{gt}\) is a binary segmentation map that distinguishes dense and sparse regions based on a threshold value \(\epsilon\). Specifically, it can be obtained with the indicator function \(\mathbbm{1}\left(y^{gt} >\epsilon\right)\). \(\alpha\) is a weight factor. \(J\) and \(\alpha\) are set to 3 and 0.01, respectively. \(\mathcal{L}_{TV}\) represents the total variation loss. We also test two commonly used \(\mathcal{L}^{s}_{reg}\) functions and the results are presented in the last section of ablation study. 

We apply the standard cross-entropy loss on labeled samples to optimize the density level classification task, which can be expressed as
\begin{equation}
    \mathcal{L}^{s}_{cls} = \frac{1}{N_l}\sum_{i}^{N_l}\mathcal{H}\left(p^{gt}_{i}, \hat{p}_{i} \right),
\end{equation}
where \(\mathcal{H(\cdot,\cdot)}\) is the cross-entropy function. 

\R{The employment of the joint regression and classification modeling strategy partially mitigates the inherent challenges posed by semi-supervised learning. When dealing with limited labeled data, the model’s ability to learn discriminative features through regression alone is 
diminished since the regression loss inherently lacks an explicit mechanism for modeling the relationships between different density levels, further compounded by the problem of overfitting to the labeled data. The inclusion of a classification task addresses this limitation by promoting the model to learn discriminative features that distinctly differentiate between different density levels. Moreover, the dual-task approach introduces an aspect of mutual regularization, thereby reducing the risk of overfitting and fostering the robustness of the feature representation. }

\R{It's also worth noting that the classification task with cross-entropy loss could diversify the feature representations. Seen through the lens of information theory, deep learning aims to maximize the mutual information between the representation and the target~\cite{shwartzziv2017opening} which means features corresponding to different targets should be well-separated, whereas features for the same target should cluster closely. However, relying solely on regression loss may not fully achieve this goal as it tends to aggregate features of the same targets without sufficiently differentiating between different targets~\cite{ZhangYMZY23}. This often results in a low-entropy feature space, which unlike in fully-supervised learning scenarios where abundant labeled data makes such compactness beneficial, poses significant challenges in semi-supervised learning. It restricts the model's ability to explore and utilize the rich structure within the unlabeled data, potentially leading to overfitting on the small labeled dataset. Moreover, it might not capture the necessary variability for robust feature learning from the unlabeled data, compromising the model's generalization ability.}

To conclude, the supervised loss can be written as 
\begin{equation}
     \mathcal{L}^{s} = \mathcal{L}^{s}_{reg} + \lambda^{s}_{cls}\cdot\mathcal{L}^{s}_{cls},
\end{equation}
where \(\lambda^{s}_{cls}\) is a weight factor for supervised classification loss.

\subsection{Unsupervised Loss}

Previous semi-supervised crowd counting methods have typically focused on performing consistency learning by selecting high-confidence predictions from unlabeled data. However, they have overlooked the issue of overfitting to local details in a counting model trained on limited labeled data, which leads to limited generalization to diverse scenes. Hence, in this proposed approach, we try to enhance model's understanding of scenes with unlabeled images by masked modeling. 

Given the majority of existing crowd counting models are designed to predict at the patch-level, we adopt patch-aligned random masking strategy~\cite{Xie00LBYD022} here as it offers the greatest flexibility in masking patches of various sizes. Each patch in an image has an equal probability of being masked and if masked, it becomes entirely invisible to the model. The student model is then presented with these masked unlabeled images and is tasked to recover the count and the density level information by leveraging the contextual cues. Meanwhile, the teacher model predicts on the fully-visible version to provide learning targets. The unsupervised loss is used for optimizing this process, which can be expressed as:
\begin{equation}
    \mathcal{L}^{u} = \mathcal{L}^{u}_{reg} + \mathcal{L}^{u}_{cls},
\end{equation}
\begin{equation}
    \mathcal{L}^{u}_{reg} = \frac{1}{N_u}\sum_{j}^{N_u}\sum_{i\in \Omega}{\mathcal{L}_1\left(\hat{y}^s_{ij}, \hat{y}^t_{ij}\right)},
\end{equation}
\begin{equation}
    \mathcal{L}^{u}_{cls} = \frac{1}{N_u}\sum_{j}^{N_u}\sum_{i\in \Omega}{\mathcal{L}_1\left(\hat{p}^s_{ij}, \hat{p}^t_{ij}\right)},
\end{equation}
where \(\Omega\) represents the collection of patches that are masked in the unlabeled images. \R{\(\hat{p}^s_{ij}\) and \(\hat{p}^t_{ij}\) represent the predicted probability distributions for sample \(i\) in \(\Omega\), produced by student and teacher model, respectively.}

Since our goal is to encourage the student network to make predictions based on holistic cues, the quality of the predictions from the teacher model is not crucial in this context. Therefore, our method avoids the necessity, as seen in most other semi-supervised crowd counting frameworks~\cite{Meng2021SpatialUS, MTCP, STC-Crowd}, of definitely requiring the establishment of a reliable uncertainty measures for regression results to filter noisy pseudo-labels. This distinction significantly increases the applicability of our framework for a wide range of crowd counting models, as obtaining uncertainty typically involves adding some complex mechanisms. Nevertheless, since no pseudo-label filtering mechanism is used in this framework, it is inevitable, particularly when dealing with datasets exhibiting substantial density diversity, that the teacher model may occasionally produce absurd predictions, potentially leading to a significant deterioration in the training process. To prevent this, we constrain the value of the regression output within a reasonable range, which is [0, 25] in our implementation.

\section{Experiments}
In this section, we begin by outlining the experimental settings, which include the crowd counting datasets and the specifics of our implementation. Subsequently, we provide a comprehensive comparison of the results obtained from our proposed method against the current state-of-the-art methods. Finally, we conduct a series of ablation studies to demonstrate the effectiveness of our method.

\subsection{Experimental Setting}
\subsubsection{Datasets} We evaluate our methods on four challenging crowd counting benchmarks: ShanghaiTechA and B~\cite{MCNN}, UCF-QNRF~\cite{qnrf} and JHU-Crowd++~\cite{sindagi2020jhu-crowd++}. Broadly speaking, ShanghaiTechA and B are two small datasets with the former containing much denser crowds than the latter. On the other hand, UCF-QNRF and JHU-Crowd++ are two large-scale datasets, both covering a diverse range of densities, making them more challenging. 

\textbf{ShanghaiTech A} consists of 482 images which were randomly collected from the internet. The annotations for each image range from 33 to 3139. The training set contains 300 images while the remaining 182 images are used for testing.

\textbf{ShanghaiTech B} contains 716 images that were collected from the crowded streets of Shanghai. This dataset has a lower density compared to ShanghaiTech A, with the number of people in an image ranging from 9 to 578. The training set contains 316 images while the remaining 400 images are used for testing.

\textbf{UCF-QNRF} is a congested dataset contains 1535 high-resolution images. It contains approximately 1.25 million point annotations with an average of 185 people per image. The training set contains 1201 images while the remaining 334 images are used for testing.

\textbf{JHU-Crowd++} is a large-scale dataset consisting of 4327 images with 1.51 million annotations. The training set contains 2272 images while the validation set has 500 images. The remaining 1600 images are used for testing.
\begin{table*}[!htbp]
\caption{Comparisons with the state-of-the-art methods on ShanghaiTech A, ShanghaiTech B, UCF-QNRF and JHU-Crowd++ datasets under \R{three} labeling ratios. The results from other methods are reported as in the original papers. The best performance is shown in \textbf{bold} while the second best is \underline{underlined}.}
\resizebox{\textwidth}{!}{
  \begin{tabular}{l|c|c|cc|cc|cc|cc}
    \toprule
    \multirow{2}{*}{Methods} & \multirow{2}{*}{Venue}& Labeled &\multicolumn{2}{c|}{ShanghaiTech A}&\multicolumn{2}{c|}{ShanghaiTech B}& \multicolumn{2}{c|}{UCF-QNRF} & \multicolumn{2}{c}{JHU-Crowd++}\\
    \cline{4-11}
    &&percentage  & MAE & MSE &  MAE & MSE & MAE & MSE & MAE & MSE \\
    \midrule
    MT~\cite{MT}&NIPS'17 & 5\% & 104.7&156.9&19.3&33.2&172.4& 284.9&101.5&363.5\\
    L2R~\cite{Liu2018cvpr}&CVPR'18& 5\% & 103.0& 155.4& 20.3& 27.6& 160.1 &272.3 &101.4 &338.8 \\
    GP~\cite{GP_count} &ECCV'20&5\% & 102.0 &172.0 &15.7& 27.9&160.0& 275.0&-&-\\
    DACount~\cite{DACount} &ACM MM'22&5\%&85.4& 134.5& \underline{12.6}& 22.8&120.2& 209.3& \underline{82.2}& \underline{294.9}\\
    OT-M~\cite{Lin_2023_CVPR}&CVPR'23&5\%&\underline{83.7}&\underline{133.3}& \underline{12.6}&\underline{21.5}&\underline{118.4}&\underline{195.4}&82.7&304.5\\
    MRC-Crowd (Ours)&- &5\%& \textbf{74.8}& \textbf{117.3}& \textbf{11.7}& \textbf{17.8}& \textbf{101.4}& \textbf{171.3}& \textbf{76.5}&\textbf{282.7}\\
    \midrule
    MT~\cite{MT}&NIPS'17 & 10\% & 94.5&156.1&15.6&24.5&145.5& 250.3&90.2&319.3\\
    L2R~\cite{Liu2018cvpr}&CVPR'18& 10\% & 90.3 &153.5& 15.6& 24.4& 148.9 &249.8 &87.5& 315.3 \\
    IRAST~\cite{liu2020eccv}&ECCV'20&10\%& 86.9& 148.9& 14.7& 22.9&-&-&-&-\\
    DACount~\cite{DACount}&ACM MM'22&10\%&74.9 &\underline{115.5}& 11.1& \underline{19.1}&\underline{109.0}& 187.2& 75.9 &282.3\\
    MTCP~\cite{MTCP}&T-NNLS'23&10\% &81.3& 130.5&14.5&22.3&124.7&206.3&88.1&318.7\\
    STC-Crowd~\cite{STC-Crowd}&T-CSVT'23&10\%&\underline{72.5} &118.2&11.7&23.8&123.7&200.9&103.1&324.7\\
    OT-M~\cite{Lin_2023_CVPR}&CVPR'23&10\%&80.1&118.5& \underline{10.8}&\textbf{18.2}&113.1&\underline{186.7}&\underline{73.0}&\underline{280.6}\\
    MRC-Crowd (Ours)&-&10\%& \textbf{67.3}& \textbf{106.8}& \textbf{10.3}& \textbf{18.2}& \textbf{93.4}& \textbf{153.2}& \textbf{70.7}& \textbf{261.3}\\
    \midrule
    MT~\cite{MT}&NIPS'17 & 40\% & 88.2&151.1&15.9&25.7&147.2& 247.6&121.5&388.9\\
    L2R~\cite{Liu2018cvpr}&CVPR'18& 40\% & 86.5 &148.2 &16.8 &25.1& 145.1 &256.1 &123.6& 376.1\\
    GP~\cite{GP_count}&ECCV'20&40\%& 89.0 & 148.0& -& - & 136.0& 218.0&-&-\\
    SUA~\cite{Meng2021SpatialUS}&ICCV'21&40\%&68.5 &121.9& 14.1& 20.6&130.3 &226.3 &80.7 &290.8\\
    DACount~\cite{DACount}&ACM MM'22&40\%&67.5 &110.7& 9.6& 14.6&\underline{91.1} &\underline{153.4}& \underline{65.1}& \underline{260.0}\\
    STC-Crowd~\cite{STC-Crowd}&T-CSVT'23&40\%&\underline{65.1}&\underline{106.3}&9.8&16.5&98.7&175.6&72.8&280.4\\
    OT-M~\cite{Lin_2023_CVPR}&CVPR'23&40\%&70.7&114.5& \underline{8.1}&\textbf{13.1}&100.6&167.6&72.1&272.0\\
    MRC-Crowd (Ours)&-&40\%& \textbf{62.1}& \textbf{95.5}& \textbf{7.8}& \underline{13.3}& \textbf{81.1}& \textbf{131.5}& \textbf{60.0}&\textbf{227.3}\\
  \bottomrule
 \end{tabular}}
\label{tab:overall_results}
\end{table*}

\subsubsection{Implementation Details} The ground-truth density maps for images in datasets other than ShanghaiTech B  are generated by using geometry adaptive kernels as described by Li et al.~\cite{li2018csrnet}. For images in ShanghaiTech B, we use the fixed Gaussian kernel with a kernel size of 4 to generate the ground-truth density maps. Furthermore, we constrain the longer side of each image within 1920 pixels in all datasets. 

VGG-19 is utilized as the backbone model, with the subsequent layers after the last pooling layer removed. We initialize it with the weights that has been pretrained on the ImageNet. We adopt the method introduced by Wang \textit{et al}.~\cite{WangSZWTHWL0W21} to partition the intervals and the number of density levels is set as 25. Please refer to their paper for a detailed ablation on this setting. AdamW~\cite{Loshchilov2017DecoupledWD} is used for optimzing the model with a learning rate of \(10^{-5}\). We apply random cropping for each training image. The crop size is 256\(\times\) 256 for ShanghaiTech A, while 512 \(\times\) 512 for others. On each dataset, we test three popular labeled ratios: 5\%, 10\%, and 40\% . To ensure a fair comparison with previous works, we adopt the same experimental protocol (the same set of labeled data) as used in DACount~\cite{DACount} and OT-M~\cite{Lin_2023_CVPR}. It is worth mentioning that our backbone is also identical to theirs. The batch size is set as 8 \R{in all experiments}. For the 5\% and 10\% labeled ratio, we set the ratio of labeled to unlabeled data as 1:3, while for the 40\% labeled ratio, we use a 1:1 ratio of labeled to unlabeled data. This setting is to reduce the number of iterations in each epoch. The masked patch size and the masking ratio are set as 32 and 0.3, respectively. Regards to the weight factors, \(\lambda_u\) and \(\lambda^{s}_{cls}\) are both set as 1 in all experiments. The weight of \(\lambda_u\) ramps up from 0 to its final value during the first 20 epochs. 

We evaluate our method with two metrics: mean absolute error (MAE) and mean squared error (MSE), which can be expressed as:

\begin{equation}
    MAE = \sum_{i=1}^{N}\mathopen|\hat{C}_i - C_i^{gt}\mathclose|,
\end{equation}
\begin{equation}
        \label{RMSE}
        MSE = \sqrt{\frac{1}{N} \sum_{i=1}^{N}(\hat{C}_i - C_i^{gt})^2},
\end{equation}
where \(N\) is the total number of images and \(\hat{C}_i\) and \(C_i^{gt}\) are the predicted and ground-truth count of \(i\)-th image, respectively.

\begin{figure*}[t]
    \centering
    \includegraphics[width=\linewidth]{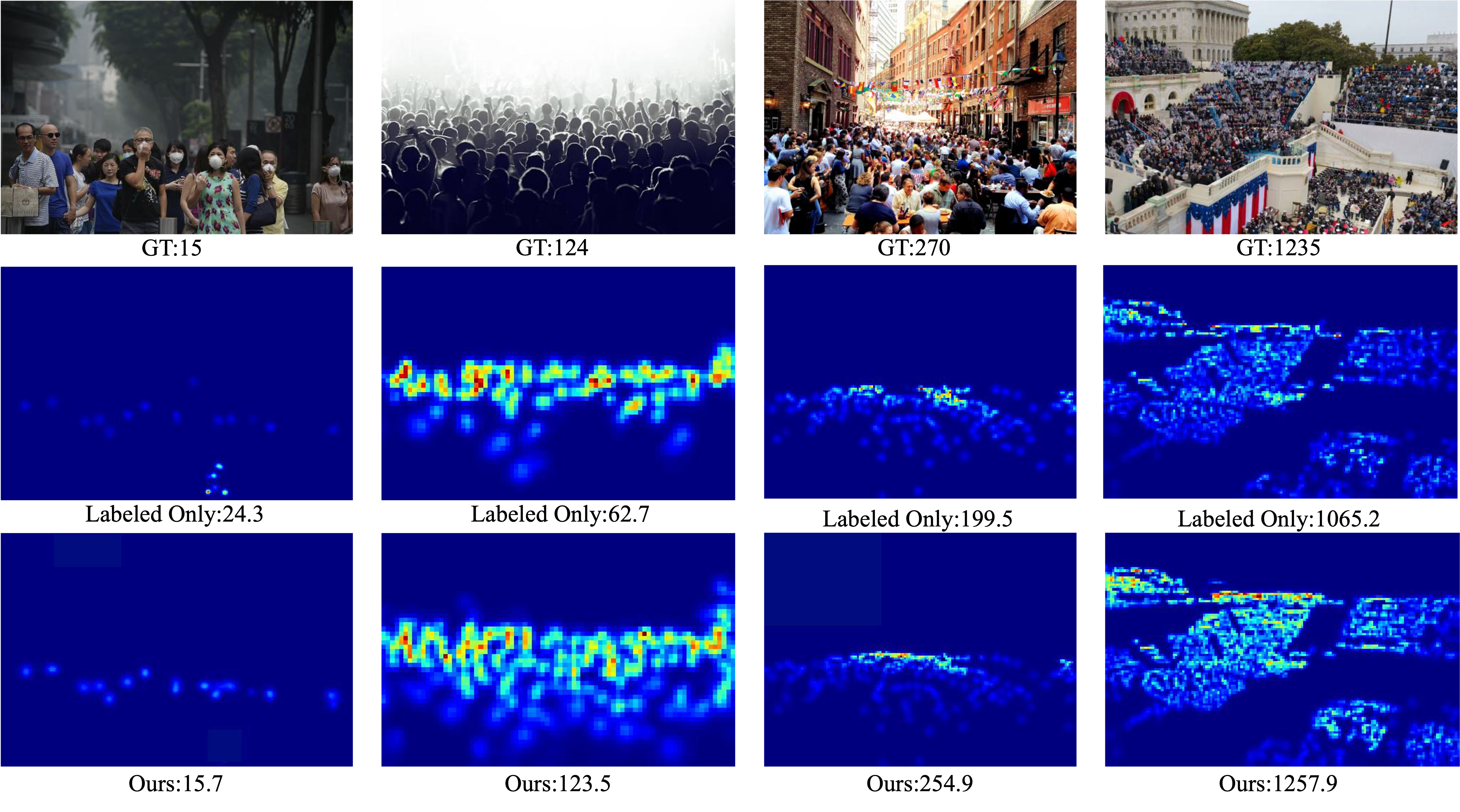}
    \caption{Visualizations of the results on the testing data of JHU-Crowd++ datasets. The first row is the input images. The second row is the output of the base model trained solely with labeled data, while the third row is the output of our proposed MRC-Crowd. Both models are trained with a labeled ratio of 5\%. }
    \label{fig:vis}
\end{figure*}

\subsection{Results}
We assess the performance of our method on the aforementioned datasets and compare its performance to the previous state-of-the-art approaches. The results are summarized in Table~\ref{tab:overall_results}. \R{A visualization of typical results on JHU-Crowd++ under a 5\% partition protocol is shown in Fig~\ref{fig:vis}. }

It is clear that our method outperforms the previous state-of-the-art methods by a large margin across different benchmarks under different values of the labeled ratio. To be specific, on ShanghaiTech A, our method reduces the MAE and MSE by an average of 8.0\% and 10.0\%, respectively, compared to the second-best method, across the three labeling ratios. We achieve competitive results on ShanghaiTech B. Notably, our method performs remarkably well on the challenging setting with a 5\% labeled ratio, achieving the best performance and improving the MAE and MSE by 7.1\% and 17.2\%, respectively. Our method also outperforms others on large-scale datasets. On UCF-QNRF, MRC-Crowd achieves a significant reduction in MAE and MSE by over 13.2\% and 14.8\% on average across three labeling ratios. In addition, our method's performance on JHU-Crowd++ further demonstrates its advantage over previous approaches which achieves the best MAE and MSE in all three settings.

\subsection{Ablation Study}

We conduct extensive ablation experiments to validate the effectiveness of our proposed framework. We first discuss the benefits of learning with unlabeled images, followed by an exploration of hyper-parameters. In the end, we show the generalizability of the proposed framework.
\subsubsection{The influence of unlabeled images}

We conduct experiments on ShanghaiTech A and UCF-QNRF to study the influence of unlabeled data on the model under all three settings. To be specific, training with only labeled data is equivalent to a conventional fully-supervised learning using \(\mathcal{L}^{s}\) without using our framework. While, our framework trains the model on both labeled and unlabeled with the combination of loss  \(\mathcal{L}^{s}\) and \(\lambda_u \cdot\mathcal{L}^{u}\). The result is presented in Table~\ref{tab:unlabelled}.  

With the help of unlabeled data and our proposed framework, we observe a significant improvement in the performance of the model. Especially on the challenging UCF-QNRF dataset, we achieve a remarkable reduction in MAE and MSE of over 19\% and 17\%, respectively, under all three settings. This quantitative results demonstrate the effectiveness of our method in utilizing unlabeled data.

\begin{table*}[t]
\centering
\caption{ The results of ablation study on the impact of unlabeled images conducted on the ShanghaiTech A and UCF-QNRF.}

\begin{tabular}{l|c|cc|cc} 
\toprule
\multirow{2}{*}{Loss} & \multirow{2}{*}{Labeled Percentage}& 
\multicolumn{2}{c|}{ShanghaiTech A} &\multicolumn{2}{c}{UCF-QNRF} \\
\cline{3-6}
&& MAE & MSE & MAE & MSE \\
\hline  
\(\mathcal{L}^{s}\) & 5\%& 86.3& 131.6 & 128.3& 208.5 \\
\(\mathcal{L}^{s} + \lambda_u \cdot\mathcal{L}^{u}\)  & 5\%& 74.8 & 117.3 & 101.4 & 171.3 \\\hline
\(\mathcal{L}^{s}\) & 10\%& 78.5 & 126.3 &115.6&184.9\\
\(\mathcal{L}^{s} + \lambda_u \cdot\mathcal{L}^{u}\)  & 10\%& 67.3 & 106.8 &93.4&153.2\\\hline
\(\mathcal{L}^{s}\) & 40\%& 70.6 & 110.2 &98.9&168.6\\
\(\mathcal{L}^{s} + \lambda_u \cdot\mathcal{L}^{u}\)  & 40\%& 62.1 & 95.5 &81.1&131.5\\
\bottomrule
\end{tabular}
\label{tab:unlabelled}
\end{table*}

\subsubsection{The impact of classification head}
\begin{table}[t]
\caption{ Ablation study on the impact of classification head conducted on ShanghaiTech A under the 5\% labeled ratio setting. }
\centering
\begin{tabular}{l|cc} 
\toprule
Loss & MAE & MSE\\
\hline  
\(\mathcal{L}^{s}_{reg}\) & 86.8& 139.2 \\
\(\mathcal{L}^{s}_{reg} + \lambda^{s}_{cls}\cdot\mathcal{L}^{s}_{cls}\) & 86.3  & 131.6  \\
\(\mathcal{L}^{s} +  \lambda_u \cdot\mathcal{L}^{u}_{reg}\) & 83.5 &129.1 \\
\(\mathcal{L}^{s} +  \lambda_u \cdot\mathcal{L}^{u}_{cls}\) & 79.3 & 120.6\\
\(\mathcal{L}^{s} + \lambda_u\cdot( \mathcal{L}^{u}_{reg} + \mathcal{L}^{u}_{cls})\) & \textbf{74.8} & \textbf{117.3} \\
\bottomrule
\end{tabular}
\label{tab:cls_head}
\end{table}
\begin{table}[t]
\centering
\caption{ Ablation study on the scale of \(\lambda_u\) conducted on ShanghaiTech A under the setting of a 5\% labeled ratio..}
\begin{tabular}{l|c|c|c|c|c} 
\toprule
\(\lambda_u\) & 0.0 & 0.001 & 0.005 & 0.01 & 0.05 \\
\hline  
  MAE& 86.3 & 85.8 & 83.5 & 84.7 & 80.8 \\
  MSE& 131.6 & 130.2 & 134.5 & 131.1 & 126.3 \\
 \hline 
 \(\lambda_u\) & 0.1 & 0.5& 1.0 & 1.5 & 2.0 \\
 \hline
 MAE& 82.1 & 77.9 & \textbf{74.8} & 77.1 & 79.2 \\
 MSE& 124.8 & 122.3 & \textbf{117.3} & 121.2 & 124.9 \\
\bottomrule
\end{tabular}

\label{tab:unlabel_param}
\end{table}

To study the influence of the fine-grained density classification task in our framework, we conduct a series of experiments on ShanghaiTech A under a 5\% labeled ratio. We begin by training a model that excludes the classification task, using only labeled data with \(\mathcal{L}_{reg}^{s}\). Then, the classification module is incorporated into the model, and is trained with \(\lambda^{s}_{cls}\cdot\mathcal{L}^{s}_{cls}\). Next, we  use the proposed framework to train the model on both labeled and unlabeled data by imposing supervision on the regression head only, the classification head only, and both heads simultaneously, with losses \(\lambda_u \cdot\mathcal{L}^{u}_{reg}\), \(\lambda_u \cdot\mathcal{L}^{u}_{cls}\), and \(\lambda_u\cdot( \mathcal{L}^{u}_{reg} + \mathcal{L}^{u}_{cls})\), respectively. The comparison result is shown in Table~\ref{tab:cls_head}.

When training with only labeled data, the inclusion of the fine-grained classification task has a positive impact on the model's performance, leading to a reduction in MSE by 7.6. The benefits of the task are more prominently manifested in the learning from unlabeled data. The results demonstrate that applying consistency constraints to the classification head is more efficient than applying them solely to the regression head, leading to a further reduction of 4.2 in MAE and 8.5 in MSE. The best performance is achieved by applying constraints on both heads. Hence, we deduce that the classification task can effectively model the density relationships between different regions, thereby better exploring the underlying data distribution in manifold space compared to the regression task. 

\subsubsection{The impact of semi-parameter \(\lambda_u\)}

\begin{table}[t]
\caption{ Ablation study on the choice of masking strategy conducted on ShanghaiTech A and UCF-QNRF under the setting of a 5\% labeled ratio.}
\centering
\begin{tabular}{c|c|cc|cc} 
\toprule
Mask & Mask& 
\multicolumn{2}{c|}{ShanghaiTech A} &\multicolumn{2}{c}{UCF-QNRF} \\
\cline{3-6}
Size& Ratio& MAE & MSE & MAE & MSE \\
\hline  
\multirow{4}{*}{8} & 0.1 & 82.3 & 126.5 & 115.1 & 194.7\\
&0.3 & 82.0 & 125.8 & 116.3 & 192.5\\
&0.5 & 82.3 & 124.5 & 111.9 & 192.1\\
&0.7 & 82.5 & 123.3 & 119.9 & 207.3\\\hline
\multirow{4}{*}{16} & 0.1 & 81.4 & 128.9 & 107.6 & 177.8 \\
&0.3 & 79.9 & 129.3 & 106.2 & 178.3 \\
&0.5 & 82.9 & 123.6 & 109.0 & 183.7 \\
&0.7 & 82.2 & 127.7 & 114.5 & 191.2 \\\hline
\multirow{4}{*}{32} & 0.1 & 78.3 & 122.6 & 103.5 & 177.9 \\
&0.3 & 74.8 & \textbf{117.3} & \textbf{101.4} & \textbf{171.3}\\
&0.5 & 79.1 & 125.0 & 104.2 & 173.3 \\
&0.7 & 83.5 & 128.5 & 106.5 & 182.1 \\\hline
\multirow{4}{*}{64} & 0.1 & 77.2 & 120.8 & 106.7 & 177.2 \\
&0.3 & \textbf{74.3} & 118.7 & 103.7 & 172.6 \\
&0.5 & 75.4 & 117.4 & 104.7  & 174.4 \\
&0.7 & 84.8 & 126.6 &  110.2 & 184.7 \\\hline
\end{tabular}

\label{tab:mask}
\end{table}

\begin{table}[t]
{\caption{ \R{Ablation study on effectiveness of masking strategy conducted on ShanghaiTech A and UCF-QNRF under the setting of a 5\% labeled ratio.}}
\centering
\R{\begin{tabular}{c|cc|cc} 
\toprule
& 
\multicolumn{2}{c|}{ShanghaiTech A} &\multicolumn{2}{c}{UCF-QNRF} \\
\cline{2-5}
& MAE & MSE & MAE & MSE \\
\hline  
w/o. masking strategy  & 81.2 & 130.5 & 124.3 & 200.7\\
w/. masking strategy & 74.8 & 117.3 & 101.4 & 171.3\\\hline
\end{tabular}}

\label{tab:mask_effect}}
\end{table}

\begin{table}[t]
\centering
\caption{ Ablation study on the  generalizability conducted on ShanghaiTech A under the setting of a 5\% labeled ratio. }
\begin{tabular}{l|c|c|cc} 
\toprule
Model & Labeled Ratio& \(\mathcal{L}^{s}_{reg}\) & MAE & \R{MSE}  \\
\hline  
\multirow{4}{*}{CSRNet~\cite{li2018csrnet}}& 5\% & \multirow{4}{*}{\(\mathcal{L}_2\)} & 93.5 & 142.6 \\
& 10\% &  & 73.7 & 123.1 \\
& 40\% &  & 70.4 & 111.7 \\
& 100\% &  & 68.2 & 115.0 \\\hline
\multirow{4}{*}{BL~\cite{ma2019bayesian}}& 5\% &\multirow{4}{*}{\(\mathcal{L}_{Bayes}\)} & 85.1 & 135.0 \\
& 10\% &  & 72.4 & 114.5 \\
& 40\% &  & 63.2 & 105.9 \\
& 100\% &  & 62.8  & 101.8 \\\hline
\end{tabular}
\label{tab:generalizability}
\end{table}

We empirically examine how the magnitude of \(\lambda_u\) affects the counting accuracy of the model by varying its value during training. All experiments are conducted on ShanghaiTech A with the labeled ratio of 5\%. The results are presented in table~\ref{tab:unlabel_param}. When \(\lambda_u \leq 0.01\), we observe that the performance of the model trained with both labeled and unlabeled images is nearly on par with the model trained solely with labeled data, which indicates that the unlabeled loss is too small to help with the training. It can be seen that the best performance is achieved when \(\lambda_u=1\), and as its value continues to increase beyond this point, the performance starts to drop. 

\subsubsection{The choice of the masking strategy} 
The masking strategy relates to the selection of the masked patch size and the masking ratio, which plays a particularly important role in our framework. To understand how the values of these two parameters influence the model's performance, we conduct a series of experiments on ShanghaiTech A and UCF-QNRF, under a 5\% labeled ratio. The results are shown in Table~\ref{tab:mask}. 

We first observe that whatever the masking strategy adopted for leveraging the unlabeled data is, the model consistently exhibits higher counting accuracy compared to the one trained solely on labeled data. However, we find that adopting a strategy with a large masked patch size (\textit{e}.\textit{g}. 32 or 64) leads to a better performance compared to using a small masked patch size. We deduce while employing masking technique could largely mitigate the issue of excessive reliance on local connections, a small masked patch size is not distant enough to enforce a model fully understand the scenes due to the relatively low complexity of the task. Moreover, the experimental results show that a low masking ratio should be used as the performance degrades at a large masking ratio (\textit{e}.\textit{g}. 0.7). We believe in such cases, the distance between visible patches becomes too significant, resulting in an inadequate amount of information for the model to learn effectively. To conclude, the strategy for this framework should be designed to have a large masked patch size with a small masking ratio.
\subsubsection{The effectiveness of the masking strategy}
\R{To further illustrate the significance of teaching model to leverage holistic cues, we train models under all identical experimental conditions with the sole exception being the exclusion of the masking strategy. The results are given in Table~\ref{tab:mask_effect}. We observe a significant performance gain with the implementation of the masking strategy. Specifically, on ShanghaiTechA, the MAE and MSE dropped by 7.8\% and 10.1\%, respectively. When models are fed with masked images, they are compelled to rely on holistic cues for making predictions in the masked regions, as local features are no longer accessible. This requirement encourages the model to utilize global features more effectively, thereby increasing its robustness. }

\subsubsection{The generalizability of our method} To demonstrate the generalizability of the proposed framework, we train two classical crowd counting models, CSRNet~\cite{li2018csrnet} and BL~\cite{ma2019bayesian}, using our framework under varying labeled ratio setting on ShanghaiTech A. Specifically, the structural change is made by adding a classification head right before the decoder in each model. Furthermore, \(\mathcal{L}^{s}_{reg}\) follows the original implementation of those methods, using \(\mathcal{L}_2\) loss for CSRNet and \(\mathcal{L}_{Bayes}\) loss for BL. \R{To optimize the performance, we fine-tune the weight factors to ensure a balanced contribution from each loss term. To be specific, when employing \(\mathcal{L}_2\) as the regression loss, we set \(\lambda^s_{cls}\) to 0.1. Conversely, when having \(\mathcal{L}_{Bayes}\) for regression, the weight for the regression loss is reduced to one tenth of its original value. Throughout, \(\lambda_u\) is consistently kept equal to 1.} We present the results in Table~\ref{tab:generalizability}. 

It can be seen that our method could help the model to achieve competitive results even when only 40\% of the data is labeled, as compared to the model trained on the fully labeled set. Specifically, when comparing the BL model trained with our framework to the state-of-the-art DACount~\cite{DACount}, which also employs a \(\mathcal{L}_{Bayes}\) based loss for supervising the regression task, our BL model achieves similar or even better results under the same experimental protocol. Notably, under the labeled ratio of 40\%, it leads to a further reduction in MAE by 4.3 and MSE by 4.8, respectively. It is worth norting that for fair comparison, the BL model trained here shares the same backbone structure as DACount. However, DACount leverages a complex transformer module to help modeling long-range relationships between pixels and redesigns the \(\mathcal{L}_{Bayes}\) for better performance on low-density regions under the semi-supervised setting. Remarkably, our method achieves better performance with a simpler design.

\subsection{Discussion}
\begin{figure}
    \centering
    \includegraphics[width=\linewidth]{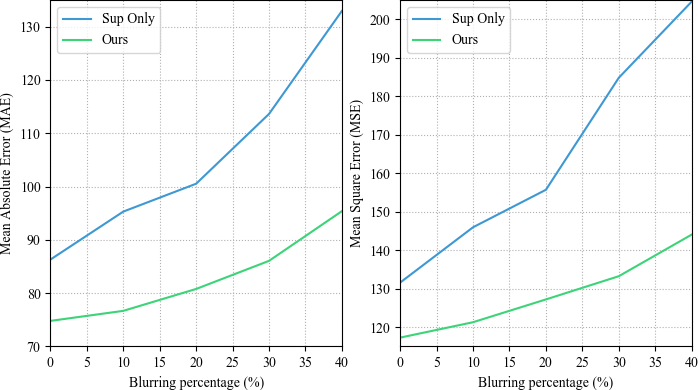}
    \caption{\R{The plots demonstrate the impact of progressively blurring patches within images on the performance of models trained exclusively with labeled data (referred to as "Sup Only") versus our proposed framework ("ours").}}
    \label{fig:blr_exp}
\end{figure}
\begin{figure}
    \centering
    \includegraphics[width=\linewidth]{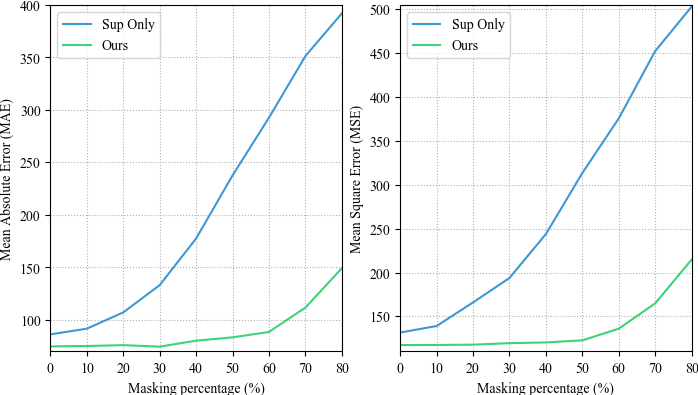}
    \caption{\R{The plots demonstrate the impact of progressively masking patches within images on the performance of models trained exclusively with labeled data (referred to as "Sup Only") versus our proposed framework ("ours").}}
    \label{fig:over_exp}
\end{figure}

\R{In this section, we conduct quantitative analysis to demonstrate the issue of overfitting to local details encountered by models when trained on limited data, and assess the contextual modeling capabilities of the proposed framework. All experiments are conducted on the test set of ShanghaiTech A and models are trained under the 5\% labeled ratio.}

\R{To address the first aim, we alter the visual information of the dataset. Specifically, we introduce a progressive blurring effect to 32 \(\times\) 32 patches across the images by adding Gaussian noise with a standard deviation of 50. The result is presented in Fig.~\ref{fig:blr_exp}. Notably, after blurring 40\% of image areas, the model trained on solely labeled data shows a significant decline in performance with both MAE and MSE increased by 54.04\% and 55.4\%, respectively. This significant deterioration in accuracy indicates the vulnerability of the model to even minor variations in visual detail. In contrast, the model trained with MRC-Crowd demonstrates a more robust performance, with a reduced decline of 27.5\% in MAE and 22.8\% in MSE, suggesting it alleviates the issue of overfitting to local detail.}

\R{Regarding the second aim, we design experiments that involve randomly masking portions of  32 \(\times\) 32 patches within images. By this means, we simulate conditions where the model must rely on indirect information to estimate crowd sizes. The objective is to assess how well the model can infer crowd counts based on the surrounding visual context. As depicted in Fig~\ref{fig:over_exp}, our framework offers strong contextual modeling capabilities, enabling the model to leverage holistic cues for making robust predictions for the masked patches. To be specific, even when the masking ratio reaches 50\%, the increases in MAE and MSE are modest, at 11.6\% and 4.5\%, respectively. Meanwhile, the model trained exclusively on labeled data exhibits a pronounced degradation in performance, with both MAE and MSE metrics doubling. As the experiment progresses with an increased proportion of masked patches, the decline in the model performance trained with MRC-Crowd becomes evident. This observation aligns with the understanding that our model's robustness and its ability to infer crowd sizes are contingent upon the availability of contextual information.}

\section{Conclusion}

Inspired by the importance of holistic patterns the cognitive phenomenon of subitizing, we introduce a simple yet powerful semi-supervised crowd counting framework called MRC-Crowd in this paper. Specifically, we propose leveraging unlabeled images to cultivate a comprehensive understanding of the crowd scenes. Our method is based on the classical mean teacher framework.  Specifically, in this framework, the student model is encouraged to predict on invisible patches of unlabeled images using holistic cues. To supervise this learning process, we leverage the predictions from the teacher model on the fully-visible images as guidance. With our framework, we observe that the trained model exhibits \R{strong contextual modeling capabilities which allows it to make robust predictions with holistic cues.} Our method imposes no restrictions, such as loss or structural requirements, on the model to be trained. It only requires a two-layer classification head as a plug-in module to enhance overall performance; however, even that module is not strictly necessary. This flexibility enables the application of our framework to a wide range of existing crowd counting models. We perform extensive experiments on four crowd benchmarks, and it could achieve the new state-of-the-art performance. 
\bibliographystyle{IEEEtran}
\bibliography{semi-main}
\vfill

\end{document}